\documentclass{article}

\usepackage{arxiv}

\usepackage[utf8]{inputenc} 
\usepackage[T1]{fontenc}    
\usepackage{hyperref}       
\usepackage{url}            
\usepackage{booktabs}       
\usepackage{amsfonts}       
\usepackage{nicefrac}       
\usepackage{microtype}      
\usepackage{lipsum}

\usepackage[utf8]{inputenc}
\usepackage{graphicx} 
\usepackage{multirow}
\usepackage{multicol} 
\usepackage{booktabs}
\usepackage{amssymb}
\usepackage{pifont}
\usepackage{textcomp}

\usepackage{caption}
\usepackage{subcaption}
\usepackage{bm}
\usepackage{amsmath}
\usepackage{mathtools}
\usepackage{commath}
\usepackage{nomencl}

\usepackage{tikz}
\usepackage{standalone}
\usepackage{bbm}
\usepackage{threeparttable}
\usepackage{hyperref}
\usepackage{cleveref}

\graphicspath{{figs/}}

\begin{document}

\title{Graph InfoClust: Leveraging cluster-level node information for unsupervised graph representation learning}

\author{
  Costas Mavromatis \\
  Department of Computer Science and Engineering\\
  University of Minnesota, USA\\
  \texttt{mavro016@umn.edu} \\
   \And
 George Karypis \\
  Department of Computer Science and Engineering\\
  University of Minnesota, USA\\
  \texttt{karypis@umn.edu} \\
}

\maketitle
\begin{abstract}
Unsupervised (or self-supervised) graph representation learning is essential to facilitate various graph data mining tasks when external supervision is unavailable. The challenge is to encode the information about the graph structure and the attributes associated with the nodes and edges into a low dimensional space.  
Most existing unsupervised methods promote similar representations across nodes that are topologically close. Recently, it was shown that leveraging additional graph-level information, e.g., information that is shared among all nodes, encourages the representations to be mindful of the global properties of the graph, which greatly improves their quality. However, in most graphs, there is significantly more structure that can be captured, e.g., nodes tend to belong to (multiple) clusters that represent structurally similar nodes. Motivated by this observation, we propose a graph representation learning method called Graph InfoClust (GIC), that seeks to additionally capture cluster-level information content. These clusters are computed by a differentiable $K$-means method and are jointly optimized by maximizing the mutual information between nodes of the same clusters. This optimization leads the node representations to capture richer information and nodal interactions, which improves their quality. 
Experiments show that GIC outperforms state-of-art methods in various downstream tasks (node classification, link prediction, and node clustering) with a 0.9\% to 6.1\% gain over the best competing approach, on average.

\end{abstract}


\thispagestyle{empty} 

\section{Introduction}

Graph structured data naturally emerge in various real-world applications. Such examples include social networks, citation networks, and biological networks. The challenge, from a data representation perspective, is to encode the high-dimensional, non-Euclidean information about the graph structure and the attributes associated with the nodes and edges into a low dimensional embedding space. The learned embeddings (a.k.a. representations) can then be used for various tasks, e.g., node classification, link prediction, community detection, and data visualization.  
In this paper, we focus on \emph{unsupervised} representation learning methods that estimate node embeddings without using any labeled data but instead employ various self-supervision approaches. These methods eliminate the need to develop task-specific graph representation models, eliminate the cost of acquiring labeled datasets, and can lead to better representations by using large unlabeled datasets.




Many self-supervision approaches ensure that nodes that are close to each other, both in terms of the graph's topology and in terms of their available features, are also close in the embedding space. This is achieved by employing a contrastive loss between pairs of nearby and pairs of distant nodes; e.g., DeepWalk~\cite{perozzi2014deepwalk}, and GraphSAGE~\cite{hamilton2017inductive}. 
Another self-supervision approach focuses on reconstructing the existing edges of the graph based on the embedding similarity of the incident nodes; e.g.,  GAE/VGAE~\cite{kipf2016variational}. Because preserving neighbor similarities is desired, many of these methods use graph neural network (GNN) encoders, that insert an additional inductive bias that nodes share similarities with their neighbors. 
%

%
%
Motivated by the fact that GNN encoders already preserve similarities between neighboring nodes, Deep Graph Infomax (DGI)~\cite{velickovic2018deep} adopts a self-supervision approach that maximizes the mutual information (MI) between the representation of each node and the \emph{global} graph representation, which is obtained by averaging the representations of the nodes in the graph. This encourages the computed node embeddings to be mindful of the global properties of the graph, which improves the representations' quality. DGI has shown to estimate superior node representations~\cite{velickovic2018deep} and is considered to be among the best unsupervised node representation learning approaches.





By maximizing the mutual information between the representation of a node and that of the global summary, DGI 
obtains node representations that capture the information content of the entire graph. However, in most graphs, there is significantly more structure that can be captured. For example, nodes tend to belong to (multiple) clusters that represent topologically near-by nodes as well as nodes have similar structural roles but are topologically distant from each other. In such cases, methods that simultaneously maximize the mutual information between the representation of a node and the summary representation of the different clusters that this node belongs to, including that of the entire graph, will allow the node representations to capture the information content of these clusters and thus encode richer structural information.

%

\begin{figure}
\begin{center}
\fbox{\scriptsize
\begin{tabular}{r@{\hspace{2pt}}lr@{\hspace{1pt}}l}
``\textbf{x}": & Global graph summary & Colors: & Labels\\
``$\bullet$":   & Cluster summaries    & SIL: & Silhouette score
\end{tabular}
}

\begin{subfigure}[t]{0.20\columnwidth}
   \centering
   \includegraphics[width=\textwidth]{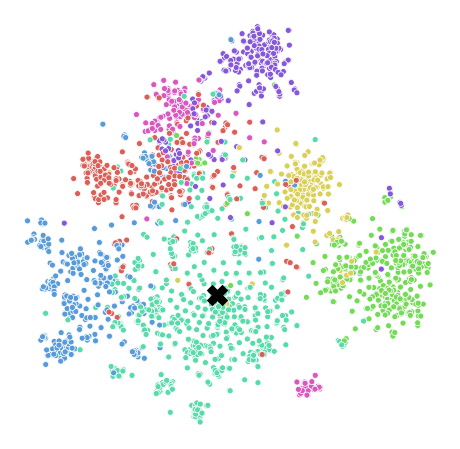}
   \caption{DGI (SIL=0.222).}
\end{subfigure}%
\hspace{15pt}
\begin{subfigure}[t]{0.20\columnwidth}
  \centering
  \includegraphics[width=\textwidth]{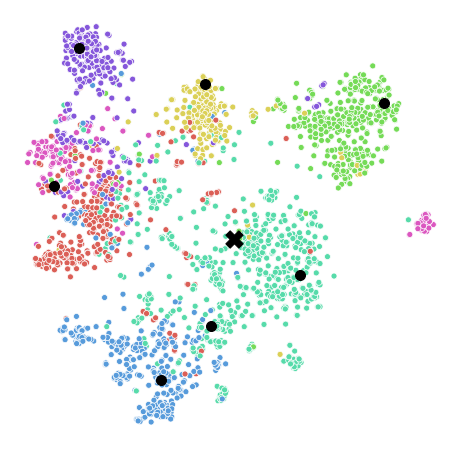}
  \caption{GIC (SIL=0.257).}
\end{subfigure}%
\end{center}
\caption{2D $t$-SNE projections~\cite{maaten2008visualizing} of the representations learned by DGI and GIC (using 7 clusters) on the CORA citation network~\cite{Mccallum00automatingthe}.}
\label{figintro}
\end{figure}

Motivated by this observation, we developed \emph{Graph InfoClust} (GIC), an unsupervised representation learning method that learns node representations by simultaneously maximizing the mutual information (MI) with respect to the graph-level summary as well as cluster-level summaries. The graph-level summary is obtained by averaging all node representations and the cluster-level summaries by a differentiable $K$-means clustering~\cite{wilder2019end} of the node representations.  The optimization of these summaries is achieved by a noise-contrastive objective, which uses discriminator functions, that discriminate between real and fake samples, as a proxy for estimating and maximizing the MI. The joint computation and optimization of the summaries, promotes both graph-level and cluster-level information and properties to the node representations, which improves their quality. 
For example, as illustrated in Fig~\ref{figintro}, GIC leads to representations that better separate the same-labeled nodes over the representations computed by DGI---the silhouette score (SIL)~\cite{rousseeuw1987silhouettes} (see Section~\ref{secexps}) of GIC is 0.257 compared to DGI's 0.222.



We evaluated GIC on seven standard datasets using node classification, link prediction, and clustering as the downstream tasks. Our experiments show that in all but two of the dataset-task combinations, GIC performs better than the best competing approach and its average improvement over DGI is 0.9, 2.6, and 15.5 percentage points for node classification, link prediction, and clustering, respectively. These results demonstrate that by 
leveraging cluster summaries, GIC is able to improve the quality of the estimated representations.

\section{Notation, Definitions, and Problem Statement}


We denote vectors by bold lower-case letters and they are assumed to be column vectors (e.g., $\mathbf{x}$). We also denote matrices by bold upper-case letters (e.g., $\mathbf{X}$). Symbol $:=$ is used in definition statements. If a matrix $\mathbf{X}$ consists of row vectors $\mathbf{x}_1, \dots, \mathbf{x}_N$, it is denoted as $\mathbf{X} := [\mathbf{x}_1, \dots, \mathbf{x}_N]$.


A graph $G$ that consists of $N$ nodes is defined as $G:=\{\mathcal{V},\mathcal{E}\}$, where $\mathcal{V} := \{n_1, \dots, n_N\}$ is the vertex set and $\mathcal{E}$ is the corresponding edge set. Connectivity is captured by the $N \times N$ adjacency matrix $\mathbf{A} \in \mathbb{R}^{N \times N}$, where $[\mathbf{A}]_{i,j} = 1 $ if $(n_i,n_j) \in \mathcal{E}$, and $[\mathbf{A}]_{i,j} = 0$ otherwise. Each node $n_i \in \mathcal{V}$ is also associated with a feature vector $\mathbf{x}_i \in \mathbb{R}^F$. All feature vectors are collected in the feature matrix $\mathbf{X} := [\mathbf{x}_1, \dots, \mathbf{x}_N]$, where $\mathbf{X} \in \mathbb{R}^{N \times F}$.


Let $\mathbf{h}_i \in \mathbb{R}^{F'}$ be an $F'$-dimensional \emph{node embedding vector} for each $n_i \in \mathcal{V}$.
Let $\mathbf{H} := [\mathbf{h}_1, \dots, \mathbf{h}_N]$, such that $\mathbf{H} \in \mathbb{R}^{N \times F'}$ is the \emph{node embedding matrix} of $G$.
The goal of \emph{node representation learning} is to learn $\mathbf{H}$ such that it preserves both $G$'s structural information $\mathbf{A}$ and its feature information $\mathbf{X}$. Once learned, $\mathbf{H}$ can be used as a single input feature matrix for downstream tasks such as node classification, link prediction, and clustering.
The problem of node representation learning is equivalent to learning an encoder function, $f: \mathbb{R}^{N \times N} \times \mathbb{R}^{N \times F} \xrightarrow{} \mathbb{R}^{N \times F'}$, that takes the adjacency matrix $\mathbf{A}$ and the feature matrix $\mathbf{X}$ as input and generates node representations  $\mathbf{H}$, namely, $ \mathbf{H} = f(\mathbf{A},\mathbf{X}) $.

\section{Related work}


\textbf{Non-GNN-based approaches}. Early unsupervised  approaches relied on matrix factorization techniques, derived by classic dimensionality reduction~\cite{belkin2003laplacian}. They use the adjacency matrix and generate embeddings by factorizing it so that similar nodes have similar embeddings, e.g., Graph factorization~\cite{ ahmed2013distributed}, GraRep ~\cite{cao2015grarep}, and HOPE ~\cite{ou2016asymmetric}.
Matrix factorization methods use deterministic measures for node similarity, thus probabilistic models were introduced to offer stochasticity. DeepWalk~\cite{perozzi2014deepwalk} and node2vec~\cite{grover2016node2vec}  optimize embeddings to encode the statistics of random walks; nodes have similar embeddings if they tend to co-occur on short random walks over the graph. Similar approaches include LINE~\cite{ tang2015line} and HARP~\cite{chen2018harp}. 
Common limitations of the aforementioned methods is that they are inherently transductive; they need to perform additional optimization rounds to generate embeddings for unseen nodes in the training phase. Generalizing machine learning models (i.e., neural networks) to the graph domain, notably with deep auto-encoders and graph neural networks (GNN), overcame these limitations and made representation learning applicable to large-scale evolving graphs. 
Deep auto-encoder approaches, such as DNGR~\cite{cao2016deep} and SDNE~\cite{wang2016structural}, use multi-layer perceptron encoders to extract node embeddings and multi-layer perceptron decoders to preserve the graph topology.

\noindent
\textbf{GNN-based approaches}. Graph neural networks (GNN) are machine learning models specifically designed for node embeddings, which operate on local neighborhoods to extract embeddings.  GNNs generate node embeddings by \textit{learning} how to  repeatedly aggregate neighbors' information. Different GNNs have been developed to promote certain graph properties~\cite{atwood2016diffusion, niepert2016learning, kipf2016semi,  hamilton2017inductive, velickovic2018graph, liu2019geniepath} and primarily differ on how they combine neighbors' information (a survey in~\cite{wu2020comprehensive}). 
As a result, successful unsupervised methods used GNN encoders to capture topology information. GAE and VGAE~\cite{kipf2016variational} use a GNN encoder to generate node embeddings and a simple decoder to reconstruct the adjacency matrix. ARGVA~\cite{pan2018adversarially} follows a similar schema with VGAE, but learns the data distribution in an adversarial manner~\cite{goodfellow2014generative}. Independently developed, GraphSAGE~\cite{hamilton2017representation} employs a GNN encoder and a random walk based objective to optimize the node embeddings.

\subsection{Deep Graph Infomax}
The aforementioned methods share the general idea that nodes that are close in the input data (graph structure or graph structure plus features), should also be as close as possible in the embedding space.
Motivated by the fact that GNNs already account for neighborhood information (i.e., are already local-biased), the work of Deep Graph Infomax (DGI)~\cite{velickovic2018deep} (and inspirational to us) employs an alternative loss function that encourages node embeddings to be mindful of the global structural properties.

The basic idea is to train the GNN-encoder $f_{GNN}$ to maximize the mutual information (MI)~\cite{shannon1948mathematical, cover1991elements} between node (fine-grain) representations, i.e., $\mathbf{H} = f_{GNN}(\mathbf{A},\mathbf{X})$, and a global representation (summary of all representations).  This encourages the encoder to prefer the information that is shared across all nodes. If some specific information, e.g., noise, is present in some neighborhoods only, this information would not increase the MI and thus, would not be preferred to be encoded. 


Maximizing the precise value of mutual information is intractable, thus, a Jensen-Shannon MI estimator is often used~\cite{hjelm2018learning, oord2018representation}, which maximizes MI's lower bound. The Jensen-Shannon-based estimator acts like a standard binary cross-entropy (BCE) loss, whose objective maximizes the expected $\log$-ratio of the samples from the joint distribution (positive examples) and the product of marginal distributions (negative examples). The positive examples are pairings of $\mathbf{s}$ with $\mathbf{h}_i$ of the real input graph  $ \mathcal{G} := (\mathbf{A},\mathbf{X})$, but the negatives  are pairings of $\mathbf{s}$ with $\tilde{\mathbf{h}}_i$, which are obtained from a fake/corrupted input graph $ \tilde{\mathcal{G}} :=(\tilde{\mathbf{A}},\tilde{\mathbf{X}})$ with $\tilde{\mathbf{H}} = f_{GNN}(\tilde{\mathbf{A}},\tilde{\mathbf{X}})$. Then, a discriminator $\mathcal{D}_1: \mathbb{R}^{F'} \times \mathbb{R}^{F'} \to \mathbb{R}$ is used to assign higher scores to the positive examples than the negatives, as in~\cite{hjelm2018learning, oord2018representation}.

The Jensen-Shannon-based BCE objective is expressed as
\begin{equation}
    \begin{split}
         \mathcal{L}_1 & = \sum_{i=1}^N \mathbb{E}_{(\mathbf{X}, \mathbf{A})} \Big[ \log \mathcal{D}_1(\mathbf{h}_i, \mathbf{s}) \Big] \\
        & + \sum_{i=1}^N \mathbb{E}_{(\tilde{\mathbf{X}}, \tilde{\mathbf{A}})} \Big[ \log \big(1 - \mathcal{D}_1(\tilde{\mathbf{h}}_i, \mathbf{s}) \big) \Big]  \ ,\\
    \end{split}
    \label{eql1}
\end{equation}
with $\tilde{\mathbf{A}} \in \mathbb{R}^{N \times N}$ and $\tilde{\mathbf{X}} \in \mathbb{R}^{N \times F}$, for simplicity.  

This objective is DGI's main contribution, which leads to superior node representations~\cite{velickovic2018deep}. As a result, DGI is considered to be among the best unsupervised node representation learning approaches. 

\begin{figure}
\begin{center}
\fbox{\scriptsize
\begin{tabular}{r@{\hspace{2pt}}lr@{\hspace{1pt}}l}
``\textbf{x}": & Global graph summary & Colors: & Labels\\
``$\bullet$":   & Cluster summaries    & SIL: & Silhouette score
\end{tabular}
}

        \begin{subfigure}[t]{0.2\columnwidth}
                \centering
                ($F'=16$)
                \includegraphics[width=\textwidth]{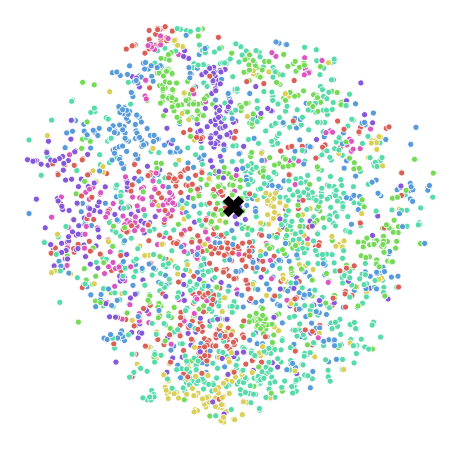}
                SIL= -0.121
        \end{subfigure}%
        \begin{subfigure}[t]{0.2\columnwidth}
                \centering
                ($F'=32$)
                \includegraphics[width=\textwidth]{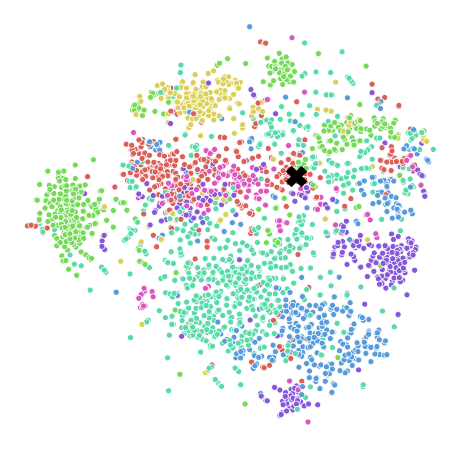}
                SIL= -0.012
        \end{subfigure}%
        \begin{subfigure}[t]{0.2\columnwidth}
                \centering
                ($F'=512$)
                \includegraphics[width=\textwidth]{motivation/dgi_512.png}
                SIL= 0.222
        \end{subfigure}%
        
        \begin{subfigure}[t]{0.2\columnwidth}
                \centering
                \includegraphics[width=\textwidth]{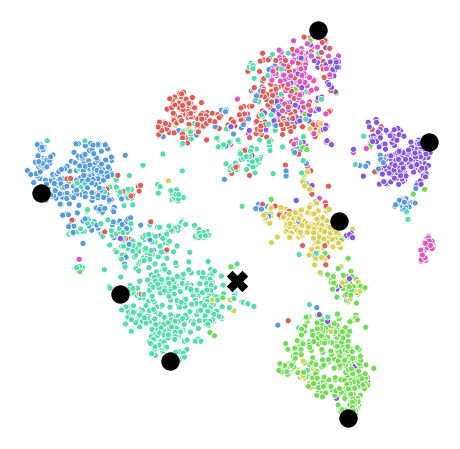}
                SIL= 0.195
        \end{subfigure}%
        \begin{subfigure}[t]{0.2\columnwidth}
                \centering
                \includegraphics[width=\textwidth]{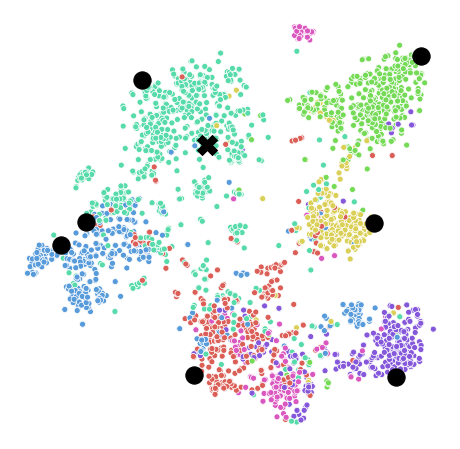}
                SIL= 0.229
        \end{subfigure}%
        \begin{subfigure}[t]{0.2\columnwidth}
                \centering
                \includegraphics[width=\textwidth]{motivation/gic_512_contr.png}
                SIL= 0.257
        \end{subfigure}%
\end{center}
\caption{Node representations (after t-SNE 2D projection is applied) and corresponding silhouette scores (SIL) (the higher the better) of leveraging a single graph summary (DGI, top) versus 7 additional cluster summaries (GIC, bottom). GIC is less sensitive to the value of the embedding dimensions $F'$ compared to DGI.}            
\label{figmotiv}
\end{figure}

\section{Graph InfoClust (GIC)}

Graph InfoClust relies on a framework similar to DGI's to optimize the embedding space so that it contains additional cluster-level information content. The novel idea is to learn node representations by maximizing the mutual information between (i) node (fine-grain) representations and the global graph summary, and (ii) node (fine-grain) representations and corresponding cluster (coarse-grain) summaries. This enables the embeddings to be mindful of various structural properties and avoids the pitfall of optimizing the embeddings based on a single vector. We explain this with Figure~\ref{figmotiv}, and show that GIC leads to better representations than DGI, especially when we limit the dimensions $F'$ of the embeddings (its capacity), and thus, the amount of information that can be encoded.

\subsection{Overview of GIC}

\begin{figure}[t!]
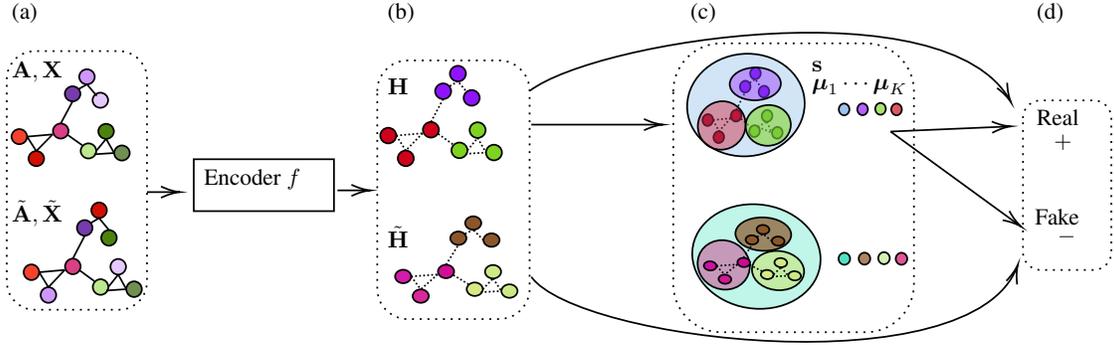

    \centering
    \includestandalone[width=.9\linewidth]{figs/framework}
    \caption{GIC's framework. (a) A fake input is created based on the real one. (b) Embeddings are computed for both inputs with a GNN-encoder. (c) The graph and cluster summaries are computed. (d) The goal is to discriminate between real and fake samples based on the computed summaries. }
    \label{figframe}
\end{figure}

GIC's overall framework is illustrated in Figure~\ref{figframe}. Mutual information is estimated and maximized through discriminator functions that discriminate between positive samples from a real input and negative samples from a fake input $\tilde{\mathcal{G}}$ (Figure~\ref{figframe}a), as in~\cite{velickovic2018deep, hjelm2018learning, oord2018representation}.
Real node embeddings $\mathbf{H}$ and fake node embeddings $\tilde{\mathbf{H}}$ are obtained by using a graph neural network (GNN) encoder $f_{GNN}$ as $\mathbf{H} = f_{GNN}(\mathbf{A},\mathbf{X})$ and $\tilde{\mathbf{H}} = f_{GNN}(\tilde{\mathbf{A}},\tilde{\mathbf{X}})$, respectively (Figure~\ref{figframe}b).

The global graph summary $\mathbf{s} \in \mathbb{R}^{1 \times F'}$ is obtained by averaging all nodes' representations, as in DGI~\cite{velickovic2018deep}. 
Cluster summaries $\boldsymbol{\mu}_k$ are obtained by, first, clustering fine-grain representations and, then, computing their summary (average of all nodes within a cluster). Suppose we want to optimize $K$ cluster summaries, so that $\boldsymbol{\mu}_k \in \mathbb{R}^{1 \times F'}$ with $k =1, \dots, K$ (Figure~\ref{figframe}c). 

The optimization is achieved by maximizing the mutual information (MI) between nodes within a cluster. In order to estimate and maximize the MI, we compute $\mathbf{z}_i \in \mathbb{R}^{1 \times F'}$ which represents the \emph{corresponding cluster summary} of each node $n_i$, based on the cluster it belongs to. Then, we can simply maximize the MI between $\mathbf{h}_i$ and $\mathbf{z}_i$ of each node. A discriminator $\mathcal{D}_K: \mathbb{R}^{F'} \times \mathbb{R}^{F'} \to \mathbb{R}$ is used as a proxy for estimating the MI by assigning higher scores to positive examples than negatives, as in~\cite{velickovic2018deep, hjelm2018learning}. We obtain positive examples by pairing $\mathbf{h}_i$ with $\mathbf{z}_i$ from the real graph and negatives by pairing $\tilde{\mathbf{h}}_i$ with $\mathbf{z}_i$ from the fake graph. The proposed objective term is given by
\begin{equation}
    \begin{split}
         \mathcal{L}_K &= \sum_{i=1}^N \mathbb{E}_{(\mathbf{X}, \mathbf{A})} \Big[ \log \mathcal{D}_K(\mathbf{h}_i, \mathbf{z}_i) \Big] \\
        & + \sum_{i=1}^N \mathbb{E}_{(\tilde{\mathbf{X}}, \tilde{\mathbf{A}})} \Big[ \log \big(1 - \mathcal{D}_K(\tilde{\mathbf{h}}_i, \mathbf{z}_i \big) \Big] \\
    \end{split}
    \label{eqlk}
\end{equation}
and GIC's overall objective is given by 
\begin{equation}
    \mathcal{L} = \alpha \mathcal{L}_1 + (1-\alpha) \mathcal{L}_K,
    \label{eqloss}
\end{equation}
where $\alpha \in [0,1]$ controls the relative importance of each component.

\subsection{Implementation Details}

\textbf{Fake input}. When the input is a single graph, we opt to corrupt the graph by row-shuffling the original features $\mathbf{X}$  as $\tilde{\mathbf{X}} := \text{shuffle}([\mathbf{x}_1, \mathbf{x}_2, \dots, \mathbf{x}_N  ])$ and $\tilde{\mathbf{A}} := \mathbf{A}$ as proposed by~\cite{velickovic2018deep} (see Figure~\ref{figframe}a). In the case of multiple input  graphs, it may be useful to randomly sample a different graph from the training set as negative examples~\cite{velickovic2018deep}.

\noindent
\textbf{Cluster and graph summaries}. The graph's summary $\mathbf{s}$ in Eq.~\eqref{eql1} (and thus in Eq.~\eqref{eqloss}), is computed as
\begin{equation}
    \mathbf{s} = \sigma\left(\frac{1}{N} \sum_{i=1}^N \mathbf{h}_i\right),
    \label{eqs}
\end{equation}
and is essentially the average of all node representations with a nonlinearity $\sigma(\cdot)$; here is the logistic sigmoid, which worked better in our case and in~\cite{velickovic2018deep}.

In Eq.~\eqref{eqlk}, in order to compute $\mathbf{z}_i$ for each node $n_i$, we apply a weighted average of the summaries of the  clusters to which node $n_i$ belongs to, as
\begin{equation}
\mathbf{z}_i = \sigma \left(\sum_{k=1}^K r_{ik} \boldsymbol{\mu}_k\right),
\label{eqzn}
\end{equation}
where $r_{ik}$ is the degree that node $n_i$ is assigned to cluster $k$, and is a soft-assignment value  (i.e., $\sum_k r_{ik} = 1, \forall i$), and $\sigma(\cdot)$ is the logistic sigmoid nonlinearity. 

The cluster summaries $\boldsymbol{\mu}_k$, with $k = 1, \dots, K$, are obtained by a layer that implements a differentiable version of $K$-means clustering, as in  ClusterNet~\cite{wilder2019end}. The ClusterNet layer updates the cluster centers in an end-to-end differentiable manner based on an input objective term, e.g., cluster modularity maximization. Here, we use Eq.~\eqref{eqlk} to optimize the clusters, and essentially maximize the intra-cluster mutual information. 
More precisely, the cluster centroids $\boldsymbol{\mu}_k$ are updated by optimizing Eq.~\eqref{eqlk} via an iterative process by alternately setting
\begin{equation}
    \boldsymbol{\mu}_k = \frac{\sum_i r_{ik} \mathbf{h}_i}{\sum_i r_{ik}} \quad   k = 1, \dots, K
    \label{eqmuk}
\end{equation}
and
\begin{equation}
    r_{ik} = \frac{\exp( -\beta \text{ sim}(\mathbf{h}_i, \boldsymbol{\mu}_k))}{\sum_k \exp( -\beta \text{ sim}(\mathbf{h}_i, \boldsymbol{\mu}_k))} \quad   k = 1, \dots, K, 
    \label{eqrnk}
\end{equation}
where $\text{sim}(\cdot, \cdot)$ denotes a similarity function between two instances and $\beta$ is an inverse-temperature hyperparameter; $\beta \to \infty$ gives a binary value for each cluster assignment. All $\mathbf{h}_i, \mathbf{z}_i$, and $r_{ik}$ for each node $n_i$ , and thus $\boldsymbol{\mu}_k$, are jointly optimized and the process is analogous to the typical $K$-means updates. It was shown that the approximate gradient with respect to the cluster centers can be calculated by unrolling a single (the last one) iteration of the forward-pass updates~\cite{wilder2019end}. This enables the final cluster output to be computed in an end-to-end fashion.

\noindent
\textbf{Discriminators}.
As the discriminator function $\mathcal{D}_1$, which is a proxy for estimating the MI between node representations and the graph summary, we use a bilinear scoring function, as proposed in~\cite{velickovic2018deep}, followed by a logistic sigmoid nonlinearity, which convert scores into probabilities, as
\begin{equation}
    \mathcal{D}_1(\mathbf{h}_i, \mathbf{s}) = \sigma(\mathbf{h}_i^T \mathbf{W} \mathbf{s}),
\end{equation}
with $\mathbf{W}$ being a learnable scoring matrix.

Moreover, we use an inner product similarity, followed by a logistic sigmoid nonlinearity $\sigma(\cdot)$, as the discriminator function $\mathcal{D}_K$ for estimating the MI between node representations and their cluster summaries as
\begin{equation}
    \mathcal{D}_K(\mathbf{h}_i, \mathbf{z}_i) = \sigma(\mathbf{h}_i^T\mathbf{z}_i) \ .
\end{equation}
Here, we replace the bilinear scoring function used for $\mathcal{D}_1$ by an inner product, since it dramatically reduces the memory requirements and worked better in our case.

\section{Experimental Methodology} \label{secexps}
\subsection{Datasets}
We evaluated the performance of GIC using seven commonly used benchmark datasets, whose statistics can be found in the Supplementary Material. Briefly, CORA~\cite{Mccallum00automatingthe}, CiteSeer~\cite{Giles98CiteSeeran}, and PubMed~\cite{namata2012query} are three citation networks, CoauthorCS and CoauthorPhysics~\cite{shchur2018pitfalls} are co-authorship graphs, and AmazonComputer and AmazonPhoto~\cite{shchur2018pitfalls} are segments of the Amazon co-purchase graph~\cite{mcauley2015image}.


\subsection{Node classification}
The goal is to predict some (or all) of the nodes' labels based on a given (small) training set. In unsupervised methods, the learned node embeddings are passed to a downstream classifier, e.g., logistic regression. Following~\cite{shchur2018pitfalls}, we sample 20$\times$\#classes nodes as the train set, 30$\times$\#classes nodes as the validation set, and the remaining nodes are the test set. The sets are either uniformly drawn from each class (balanced sets) or randomly sampled (imbalanced sets). For the unsupervised methods, we use a logistic regression classifier, which is trained with a learning rate of 0.01 for 1k epochs with Adam SGD optimizer \cite{kingma2014adam} and Glorot initialization \cite{glorot10a}. The classification accuracy (Acc) is reported as a performance metric, which is the percentage of correctly classified instances (TP + TN)/(TP + TN + FP + FN) where TP, FN, FP, and TN represent the number of true positives, false negatives, false positives, and true negatives, respectively. The final results are averaged over 20 times. Following~\cite{shchur2018pitfalls}, we set the embedding dimensions $F'=64$. 

\subsection{Link prediction}
In link prediction, some edges are hidden in the input graph and the goal is to predict the existence of these edges based on the computed embeddings. The probability of an edge between nodes $i$ and $j$ is given by $\sigma(\mathbf{h}_i^T \mathbf{h}_j) $, where $\sigma$ is the logistic sigmoid function. We follow the setup described in~\cite{kipf2016variational, pan2018adversarially}: 5\% of edges and negative edges as validation set, 10\% of edges and negative edges as test set, $F'=16$, and the results are averaged over 10 runs. We report the area under the ROC curve (AUC) score~\cite{bradley1997roc}, which is equal to the probability that a randomly chosen edge is ranked higher than a randomly chosen negative edge, and the average precision (AP) score~\cite{su2015ap}, which is the area under the precision-recall curve;  here, precision is given by TP/(TP+FP) and recall by  TP/(TP+FN).

\subsection{Clustering}
In clustering, the goal is to cluster together related nodes (e.g., nodes that belong to the same class) without any label information. The computed embeddings are clustered into $K =$ \#classes clusters with $K$-means. The evaluation is provided by external labels, the same used for node classification.  We report the classification accuracy (Acc), normalized mutual information (NMI), and average rand index (ARI)~\cite{hubert1985comparing, manning2008information}. NMI is an information-theoretic metric, while ARI can be viewed as an accuracy metric which additionally penalizes incorrect decisions. We set $F'=32$, since most state-of-art competing approaches are invariant to the embedding size choice~\cite{pan2019learning}.

\subsection{Hyper-parameter tuning and model selection}
\textbf{Encoder}. As an encoder function $f_{\text{GNN}}$ we utilize a graph convolution network (GCN)~\cite{kipf2016semi} with the following propagation rule at layer $l$
\begin{equation}
    \mathbf{H}^{(l+1)} = \sigma\big(\hat{\mathbf{D}}^{-\frac{1}{2}} \hat{\mathbf{A}} \hat{\mathbf{D}}^{-\frac{1}{2}} \mathbf{H}^{(l)} \mathbf{\Theta}\big),
\end{equation}
where $\hat{\mathbf{A}} = \mathbf{A} + \mathbf{I}_N$ is the adjacency matrix with self-loops, $\hat{\mathbf{D}}$ is the degree matrix of $\hat{\mathbf{A}}$, i.e., $\hat{D}_{ii} = \sum_j \hat{A}_{ij} $,  $\mathbf{\Theta} \in \mathbb{R}^{F \times F'}$ is a learnable matrix, $\sigma(\cdot)$ denotes a nonlinear activation (here PReLU~\cite{he2015delving}), and $\mathbf{H}^{(0)} = \mathbf{X}$.

\noindent
\textbf{Parameter selection}. We use one-layer GCN-encoder ($l=1$), as suggested by \cite{velickovic2018deep}, and as a similarity function in Eq.~\eqref{eqrnk}, we employ the cosine similarity as suggested by \cite{wilder2019end}, and we iterate the cluster updates in Eq.~\eqref{eqmuk}, Eq.~\eqref{eqrnk} for 10 times. Since GIC's cluster updates are performed in the unit sphere (cosine similarity), we row-normalize the embeddings before the downstream task.

GIC's learnable parameters are initialized with Glorot initialization~\cite{glorot10a} and the objective is optimized using the Adam SGD optimizer~\cite{kingma2014adam} with a learning rate  of~0.001. We train for a maximum of $2k$ epochs, but the training is terminated with early stopping if the training loss does not improve in 50 consecutive epochs. The model state is reset to the one with the best (lowest) training loss.

To study the effect of the hyperparameters, which are $\alpha$, the regularization parameter of the two objective terms, $\beta$, that controls the softness of the clusters, and $K$, the number of clusters, an ablation study is first provided. Then, for each dataset-task pair, we perform model selection based on the validation set: We set $\alpha \in \{0.25, 0.5, 0.75\}$, $\beta=\{10, 100\}$ and $K \in \{32, 128\}$ to train the model, and keep the parameters' triplet that achieved the best result on the validation set (accuracy metric for node classification, AUC for link prediction, and accuracy for clustering). After that, we continue with the \emph{same} triplet for the rest runs.

\subsection{Competing approaches}
We compare the performance of GIC against fifteen unsupervised and six semi-supervised methods and variants; details can be found in the Supplementary Material.  The results for all the competing methods, except DGI, were obtained directly from \cite{shchur2018pitfalls,pan2019learning}. For DGI, we report results based on the DGI implementation of Deep Graph Library~\cite{wang2019dgl} for node classification, and the original DGI implementation for link prediction and clustering. Oftentimes, we refer to GIC with $\alpha=1$ in Eq.~\eqref{eqloss} as DGI, since it coincides with the original DGI model.

\begin{figure}[ht]
\begin{center}
\fbox{\scriptsize
\begin{tabular}{r@{\hspace{2pt}}lr@{\hspace{1pt}}l}
``\textbf{x}": & Global graph summary & Colors: & Labels\\
``$\bullet$":   & Cluster summaries    & SIL: & Silhouette score
\end{tabular}
} 
\begin{subfigure}[b]{\linewidth}
        \begin{subfigure}[b]{.49\linewidth}
        \centering
        \begin{subfigure}[b]{.25\linewidth}
        \centering
        $K=7$
        \includegraphics[width=\linewidth]{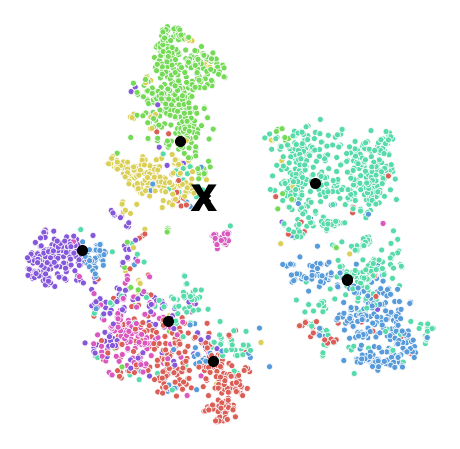}
        SIL=0.153
        \end{subfigure}
        \begin{subfigure}[b]{.25\linewidth}
        \centering
        $K=32$
        \includegraphics[width=\linewidth]{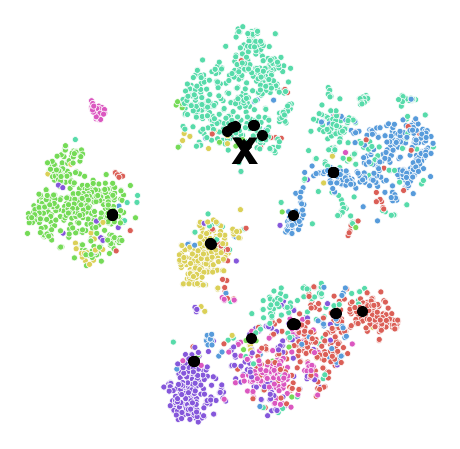}
        SIL=0.206
        \end{subfigure}
        \begin{subfigure}[b]{.25\linewidth}
        \centering
        $K=128$
        \includegraphics[width=\linewidth]{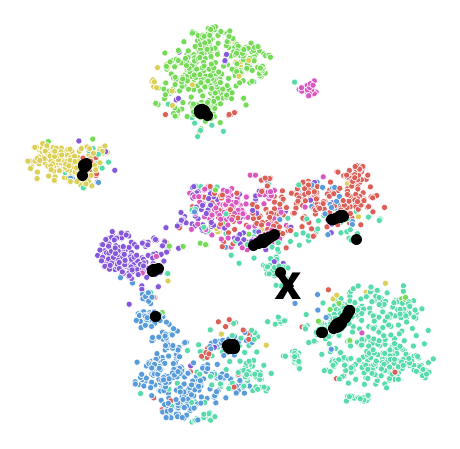}
        SIL=0.226
        \end{subfigure}
        \caption{$\alpha=0, \beta=10$}
        \label{figa0b10}
        \end{subfigure}
        \hfill
        \begin{subfigure}[b]{.49\linewidth}
        \begin{subfigure}[b]{.25\linewidth}
        \centering
        $K=7$
        \includegraphics[width=\linewidth]{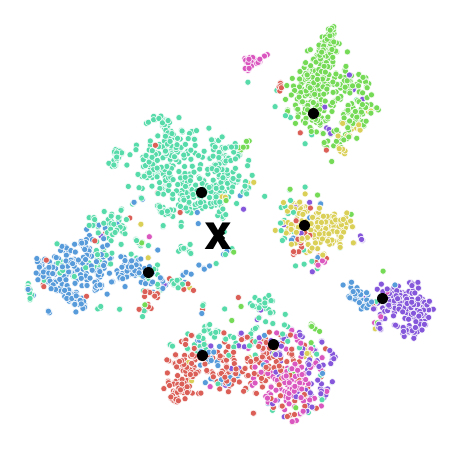}
        SIL=0.213
        \end{subfigure}
        \begin{subfigure}[b]{.25\linewidth}
        \centering
        $K=32$
        \includegraphics[width=\linewidth]{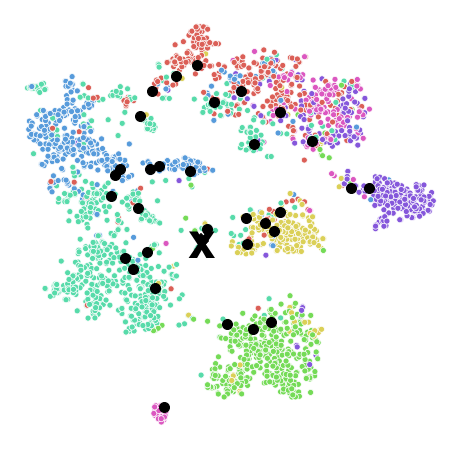}
        SIL=0.188
        \end{subfigure}
        \begin{subfigure}[b]{.25\linewidth}
        \centering
        $K=128$
        \includegraphics[width=\linewidth]{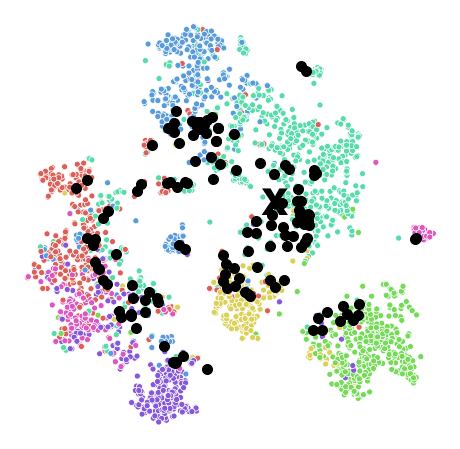}
        SIL=0.198
        \end{subfigure}
        \caption{$\alpha=0, \beta=100$}
        \label{figa0b100}
        \end{subfigure}
\end{subfigure}
\par\bigskip
\begin{subfigure}[b]{\linewidth}
    \begin{subfigure}[b]{.49\linewidth}
    \centering
        \begin{subfigure}[b]{.25\linewidth}
        \centering
        \includegraphics[width=\linewidth]{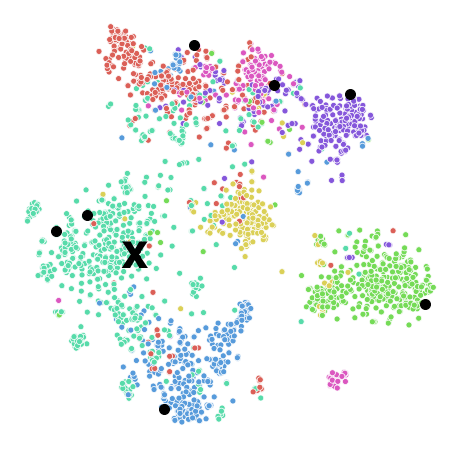}
        SIL=0.248
        \end{subfigure}
        \begin{subfigure}[b]{.25\linewidth}
        \centering
        \includegraphics[width=\linewidth]{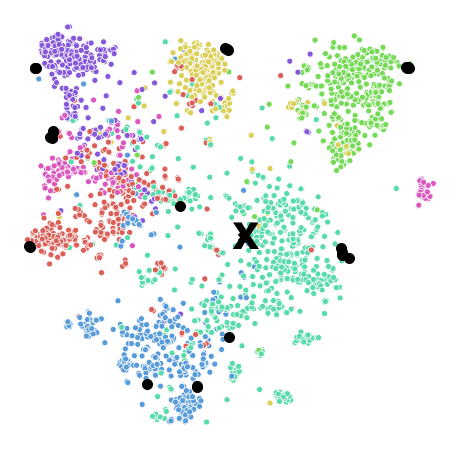}
        SIL=0.250
        \end{subfigure}
        \begin{subfigure}[b]{.25\linewidth}
        \centering
        \includegraphics[width=\linewidth]{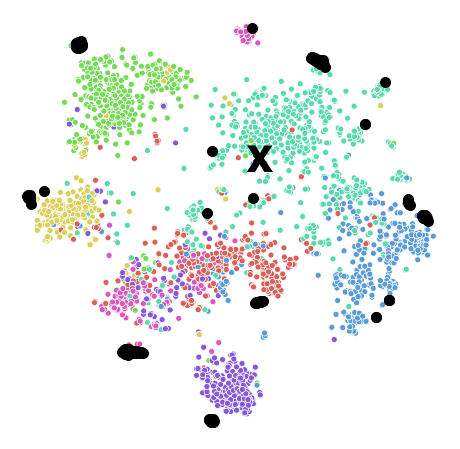}
        SIL=0.264
        \end{subfigure}
    \caption{$\alpha=0.5, \beta=10$}
    \label{figa05b10}
    \end{subfigure}
    \hfill
    \begin{subfigure}[b]{.49\linewidth}
        \begin{subfigure}[b]{.25\linewidth}
        \centering
        \includegraphics[width=\linewidth]{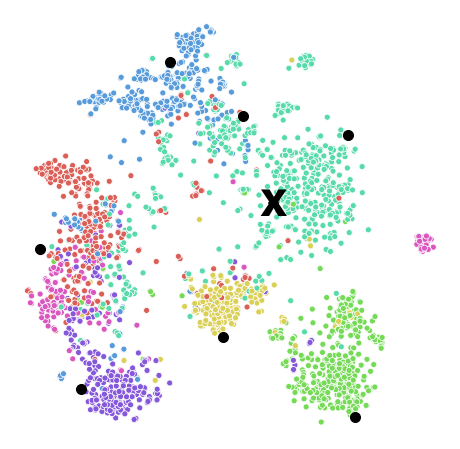}
        SIL=0.224
        \end{subfigure}
        \begin{subfigure}[b]{.25\linewidth}
        \centering
        \includegraphics[width=\linewidth]{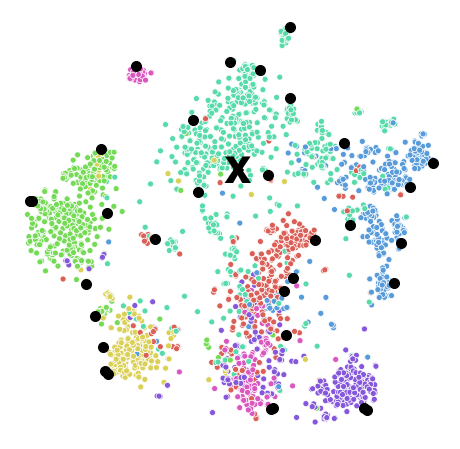}
        SIL=0.234
        \end{subfigure}
        \begin{subfigure}[b]{.25\linewidth}
        \centering
        \includegraphics[width=\linewidth]{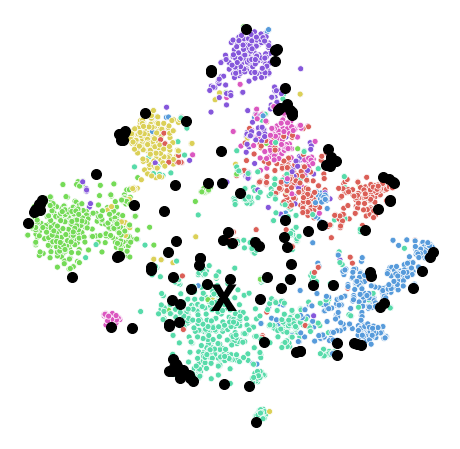}
        SIL=0.245
        \end{subfigure}
        \caption{$\alpha=0.5, \beta=100$}
        \label{figa05b100}
        \end{subfigure}
\end{subfigure}
\end{center}
\caption{t-SNE plots for CORA dataset and the corresponding silhouette scores (SIL) with $\alpha\in\{0,0.5\}, \beta \in\{10,100\}, K\in\{\#\text{classes}, 32, 128\}$. The corresponding SIL score for DGI ($\alpha=1$) is 0.212.}
\label{figablation}
\end{figure}   

\begin{table}[ht]
\small
\centering
\caption{Node classification accuracy (in \%) of various datasets and for two different train/val sets: balanced and imbalanced. The datasets are randomly split in each run.}
\label{tbnodecl}
\resizebox{0.95\textwidth}{!}{  
\begin{threeparttable}
\begin{tabular}{lllllll} 
\toprule
& \multicolumn{4}{c}{Unsupervised} & \multicolumn{2}{c}{Semi-supervised}\\
& \multicolumn{2}{c}{\textbf{GIC}} & \multicolumn{2}{c}{DGI} & Best Method &  Worst Method\\
Train/Val. Sets & Imbalanced & Balanced& Imbalanced& Balanced& \multicolumn{2}{c}{Balanced} \\
\midrule
CORA & \textbf{81.7} \textpm 1.5 & 80.7 \textpm 1.1 & 80.2 \textpm 1.8 & 80.0 \textpm 1.3 &  81.8 \textpm 1.3 (GAT) & 76.6 \textpm 1.9 (GS-maxpool) \\
 
CiteSeer & \textbf{71.9} \textpm 1.4 &  70.8 \textpm 2.0 & 71.5 \textpm 1.3 & 70.5 \textpm 1.2 &  71.9 \textpm 1.9 (GCN) & 67.5 \textpm 2.3 (GS-maxpool) \\
 
PubMed & 77.3 \textpm 1.9 & \textbf{77.4} \textpm 1.9 & 76.2 \textpm 2.0  & 76.8 \textpm 2.3 &  78.7 \textpm 2.3 (GAT)& 76.1 \textpm 2.3 (GS-maxpool) \\

 
CoauthorCS & \textbf{89.4} \textpm 0.4 &  89.3 \textpm 0.7 & 89.0 \textpm 0.4 & 88.7 \textpm 0.8 &  91.3 \textpm 2.3 (GS-mean) & 85.0 \textpm 1.1 (GS-maxpool) \\

CoauthorPhysics & \textbf{93.1} \textpm 0.7 &  92.4 \textpm 0.9 & 92.7 \textpm 0.8& 91.8 \textpm 1.0&  93.0 \textpm 0.8 (GS-mean)& 90.3 \textpm 1.2 (GS-maxpool) \\

AmazonComputers & \textbf{81.5} \textpm 1.0 &  79.5 \textpm 1.4 & 79.0 \textpm 1.7& 77.9 \textpm 1.8 &  83.5 \textpm 2.2 (MoNet)& 78.0\textpm 19.0 (GAT) \\

AmazonPhoto & \textbf{90.4} \textpm 1.0 &  89.0 \textpm 1.6 & 88.2 \textpm 1.7& 86.8 \textpm 1.7 &  91.4 \textpm 1.3 (GS-mean) & 85.7\textpm 20.3 (GAT) \\

\bottomrule
\end{tabular}
\begin{tablenotes}
\item Results are reported on the largest connected component (LCC) of the input graph.
\item Semi-supervised methods: GCN~\cite{kipf2016semi}, GAT~\cite{velickovic2018graph}, GraphSAGE (GS)~\cite{hamilton2017inductive}, and MoNet~\cite{monti2017geometric}.
\end{tablenotes}

\end{threeparttable}
}
\end{table}

\begin{table}[ht]
\centering
\caption{Link prediction scores: Area Under Curve (AUC) score and Average Precision (AP) score (in \%).  }
\label{tblink}
\resizebox{0.95\linewidth}{!}{ 
\begin{threeparttable}
\begin{tabular}{lllllll} 
\toprule
 & \multicolumn{2}{c}{\textbf{CORA}} &  \multicolumn{2}{c}{\textbf{CiteSeer}} &  \multicolumn{2}{c}{\textbf{PubMed}} \\
& AUC & AP & AUC & AP & AUC & AP\\
\midrule
Spectral Clustering~\cite{tang2011leveraging} & $84.6 \pm 0.01$ & $ 88.5 \pm 0.00$ & $ 80.5 \pm 0.01$ & $85.0 \pm 0.01$ & $84.2 \pm 0.02$ & $ 87.8 \pm 0.01$\\
DeepWalk~\cite{perozzi2014deepwalk} & $83.1 \pm 0.01$ & $ 85.0 \pm 0.00 $ & $80.5 \pm 0.02$ & $ 83.6 \pm 0.01$ & $ 84.4 \pm 0.00$ & $ 84.1 \pm 0.00$\\
VGAE~\cite{kipf2016variational} & $91.4 \pm 0.01$ &  $92.6 \pm 0.01$ &  $90.8 \pm 0.02 $ & $92.0 \pm 0.02$ & $96.4 \pm 0.00$ & $96.5 \pm 0.00$\\
ARGVA~\cite{pan2019learning} & $92.4 \pm 0.004$ & $93.2 \pm 0.003$ & $92.4 \pm 0.003$ & $93.0 \pm 0.003$ & $\textbf{96.8} \pm 0.001$ & $\textbf{97.1} \pm 0.001$\\
DGI~\cite{velickovic2018deep} & $89.8 \pm 0.8$ & $89.7 \pm 1.0$ & $95.5 \pm 1.0$ & $95.7 \pm 1.0$ & $91.2 \pm 0.6$ & $92.2 \pm 0.5$\\
\textbf{GIC} & $\textbf{93.5} \pm 0.6$ & $\textbf{93.3} \pm 0.7$ & $\textbf{97.0} \pm 0.5$ & $\textbf{96.8} \pm 0.5$ & $93.7 \pm 0.3$ & $93.5  \pm 0.3$\\
\bottomrule

\end{tabular}
\begin{tablenotes}
\item For VGAE and ARGVA, we report their best performing variant for each dataset-metric pair: four and six total variants, respectively.
\item For DGI only, we set $F'=32$ which greatly improves its results compared to $F'=16$.
\end{tablenotes}
\end{threeparttable}
}
\end{table}

\begin{table}[ht]
\centering
\caption{Clustering results with respect to the true labels.}
\label{tbcluster}
\resizebox{.85\linewidth}{!}{  
\begin{threeparttable}
\begin{tabular}{l  lll   lll   lll  } 
\toprule
 & \multicolumn{3}{c}{\textbf{CORA}} &  \multicolumn{3}{c}{\textbf{CiteSeer}} &  \multicolumn{3}{c}{\textbf{PubMed}}  \\
& Acc & NMI & ARI & Acc & NMI & ARI & Acc & NMI & ARI\\
\midrule
$K$-means & 49.2 & 31.1 & 23.0 & 54.0 & 30.5 & 27.9 & 39.8 & 0.1 & 0.2  \\
DeepWalk~\cite{perozzi2014deepwalk} & 48,4 &  32.7 & 24.3 & 33.7 & 8.8 & 9.2 & 68.4 & 27.9 & 29.9\\
DNGR~\cite{cao2016deep} & 41.9 & 31.8 &  14.2 & 32.6 & 18.0 & 4.4 & 45.8 & 15.5 & 5.4\\
TADW~\cite{yang2015network} &  56.0 & 44.1 &  33.2 & 45.5 & 29.1 & 22.8 & 35.4 & 0.1 & 0.1 \\
VGAE~\cite{kipf2016variational} & 60.9 & 43.6 & 34.7 & 40.8 & 17.6 & 12.4 & 67.2 & 27.7 & 27.9\\
ARGVA~\cite{pan2019learning} & 71.1 & 52.6 & 49.5 & 58.1 & 33.8 & 30.1 & \textbf{69.0} & 30.5 & \textbf{30.6}\\
\multirow{2}{*}{DGI~\cite{velickovic2018deep}}& $59.0$ & $38.6$ & $33.6$ & $57.9$ & $30.9$ & $27.9$ & $49.9$ & $15.1$ & $14.5$ \\

 & ($71.3$) & ($56.4$) & ($ 51.1$) & ($68.8$) & ($44.4$) & ($45.0$) & ($53.3$) & ($18.1$) & ($16.6$) \\

\textbf{GIC}& $\textbf{72.5}$ & $\textbf{53.7}$ & $\textbf{50.8}$ & $\textbf{69.6}$ & $\textbf{45.3}$ & $\textbf{46.5}$ & $67.3$& \textbf{31.9} & $29.1$\\
\bottomrule
\end{tabular}
\begin{tablenotes}
\item Acc: accuracy, NMI: normalized mutual information, ARI: average rand index in percents (\%). 
\item For VGAE and ARGVA, we report their best performing variant for each dataset-metric pair: four and six total variants, respectively.
\item For DGI, we also provide results in parentheses for $F'=512$ (CORA and CiteSeer) and $F'=128$ (PubMed), as reference.

\end{tablenotes}
\end{threeparttable}
}
\end{table} 

\section{Experimental results}

\subsection{Ablation Study} \label{secintra}
In Figure~\ref{figablation}, we plot the t-SNE 2D projection~\cite{maaten2008visualizing} of the learned node representations for the CORA dataset and $F'=64$, and use silhouette scores (SIL)~\cite{rousseeuw1987silhouettes} to evaluate the results. SIL is the the mean silhouette coefficient of all nodes, where the silhouette coefficient of each node $n_i$ is a function of (i) the average distance between $n_i$ and all other nodes with the same label and (ii) the lowest average distance of $n_i$ to all nodes in any other label. Here, these distances are computed in the 2D projected space.

As illustrated in Figure~\ref{figablation}, parameter $\alpha$ (which controls the relative importance of the two objective terms) tends to lead the optimized cluster summaries to show a certain behavior: A larger $\alpha$ pushes these summaries far from the center of the embedding space, while $\alpha=0$ usually leads them to be near the center. For example, when comparing Figure~\ref{figa05b10} ($K=128$) to Figure~\ref{figa0b10} ($K=128$), most cluster summaries are spread to the borders of the embedding space (which also achieves a better SIL score).

With alternating parameter $\beta$, the distance between two cluster summaries is affected.  A large $\beta$, e.g., $\beta=100$, makes these distances larger than a smaller one, e.g., $\beta=10$. For example, when comparing Figure~\ref{figa05b100} ($K=128$) to Figure~\ref{figa05b10} ($K=128$), for $\beta=100$, the cluster summaries are more well-located in the embedding space, while for $\beta=10$, we observe that most of them coincide. 

Although increasing $K$ generally helps, e.g., Figure~\ref{figa0b10} ($K=128$) compared to Figure~\ref{figa0b10} ($K=7$), the choice of $K$ is not independent on the choice of $\alpha, \beta$: Figure~\ref{figa0b100} is a counterexample that $K=7$ achieved the best SIL score. 

Generally, we found that setting $\alpha$ close to values of $0.5$ achieves better results $\alpha=0$ or $\alpha=1$ (see Supplementary Material). Finally, we observe the same behaviour based on $\beta$ and $K$ values in larger datasets, i.e., PubMed, as shown in the Supplementary Material.

\subsection{Comparison against competing approaches}
\textbf{Node classification}.
GIC against DGI for the task of node classification and we present the results in Table~\ref{tbnodecl}. As we can see, GIC outperforms DGI in all datasets. In CORA and PubMed, GIC achieves a mean classification accuracy gain of more than 1\%. In CiteSeer, CoauthorCS and CoauthorPhysics, the gain is slightly lower, but still more than 0.4\%, on average. In AmazonComputers and AmazonPhoto, GIC performs significantly better than DGI with a gain of more than 2\%, on average. Moreover, GIC's performance lies always in between the best and worst performing semi-supervised method. In all cases, GIC performs better than the worst performing semi-supervised method with a gain of more than 1.5\% and as high as 4.3\%. Finally, as it is later demonstrated, GIC achieves even higher gains over DGI in cases like clustering, where the downstream model does not have access to the labels. 



\noindent
\textbf{Link prediction}. Table~\ref{tblink} illustrates the benefits of GIC over DGI for link prediction tasks. GIC outperforms  DGI in all three datasets, by around 3.5\% in CORA, 1.5\% in CiteSeer, and 2.5\% in PubMed, even though GIC's embedding size is half of DGI's. GIC also outperforms VGAE and ARGVA, in CORA and CiteSeer by 1\%--2\% and 4.5\%--5.5\%, respectively. In PubMed, the performance of GIC and DGI is worse than that of VGAE and ARGVA. We believe this is because reconstruction-based methods overestimate the existing node links (they reconstruct the adjacency matrix), which favors datasets like PubMed (this is not the case for CORA and CiteSeer which have significantly more attributes to exploit). 

\noindent
\textbf{Clustering}. Table~\ref{tbcluster} illustrates GIC's performance for clustering. GIC performs better than other unsupervised methods in two out of three datasets, except for PubMed, where ARGVA works slightly better (however, we note that GIC outperforms five out of six ARGVA variants in this dataset). The gain over DGI is significantly large in all datasets, and can be as high as 15\% to 18.5\% for the NMI metric. GIC outperforms DGI in most metrics and datasets, even though DGI uses 16 times the embedding size of GIC ($F'=512$).

\section{Conclusion}
We have presented Graph InfoClust (GIC), an unsupervised graph representation learning method which relies on leveraging cluster-level content. GIC identifies nodes with similar representations, clusters them together, and maximizes their mutual information. This enables us to improve the quality of node representations with richer content and obtain better results than existing approaches for tasks like node classification, link prediction, clustering, and data visualization.


\bibliographystyle{unsrt}  
\bibliography{multiInfo}

\clearpage
\section{Supplementary Material}

\subsection{Datasets}

We evaluated the performance of GIC using seven commonly used benchmark datasets (Table~\ref{tbdata}). 

CORA, CiteSeer, and PubMed~\cite{sen2008collective,yang2016planetoid} are three text classification datasets.  Each dataset contains bag-of-words representation of documents as features and citation links between the documents as edges. The CORA dataset~\cite{Mccallum00automatingthe} contains a number of machine-learning papers divided into one of seven classes while the CiteSeer dataset~\cite{Giles98CiteSeeran} has six class labels. 
The PubMed dataset~\cite{namata2012query} consists of scientific publications from the PubMed database pertaining to diabetes classified into three classes.

CoauthorCS and CoauthorPhysics~\cite{shchur2018pitfalls} are co-authorship graphs based on the Microsoft Academic Graph. Nodes are authors and edges represent co-authorship relations. Classes represent the most active fields of studies, and node features are the bag-of-words encoded paper keywords for each author’s papers.

AmazonComputer and AmazonPhoto~\cite{shchur2018pitfalls} are segments of the Amazon co-purchase graph~\cite{mcauley2015image}. Nodes represent items and are classified into product categories, with their features to be bag-of-words encoded product reviews. Edges indicate that two items are frequently bought together.

\begin{table}
\centering
\caption{Datasets statistics; whole graph and largest connected component (LCC) of the graph as reported in~\cite{shchur2018pitfalls}. 
}
\label{tbdata}
\resizebox{.85\textwidth}{!}{  
\begin{threeparttable}
\begin{tabular}{lrrrrrrrr} 
\toprule
&Classes& Features& Nodes& Edges & Label rate & Nodes LCC & Edges LCC  & Label rate LCC\\
\midrule
CORA &7 &1,433& 2,708 & 6,632 & 0.0517 & 2,485&  5,069 & 0.0563\\
CiteSeer &6& 3,703& 3,327&  4,614& 0.0324 & 2,110 &3,668 & 0.0569\\
PubMed &3 &500 &19,717& 44,324 & 0.0030 &19,717& 44,324 &  0.0030\\
CoauthorCS &15 &6,805 &18,333  & 81,894 & 0.0164 & 18,333& 81,894 &0.0164 \\
CoauthorPhysics &5& 8,415&34,493& 247,962& 0.0029& 34,493& 247,962& 0.0029\\
AmazonComputer& 10& 767 & 13,752 & 287,209 & 0.0145 & 13,381 &245,778& 0.0149\\
AmazonPhoto& 8& 745& 7,650 & 143,663 & 0.0209 & 7,487& 119,043& 0.0214\\
\bottomrule
\end{tabular}
\begin{tablenotes}
\item Label rate is the fraction of nodes in the training set (train size equals to 20$\times$\#classes) for node classification tasks.
\end{tablenotes}

\end{threeparttable}
}
\end{table}
\subsection{Competing Approaches}

We compare the performance of GIC against the following unsupervised methods and baselines.
\begin{itemize}
    \item Deep Graph Infomax (DGI)~\cite{velickovic2018deep}
    \item Variational Graph Auto-Encoders (GAE*/VGAE* and GAE/VGAE)~\cite{kipf2016variational}
    \item Adversarially Regularized Graph Autoencoder (ARGVA)~\cite{pan2019learning}, and its variants ARGA, ARGA-DG, ARGVA-DG, ARGA-AX, and ARGVA-AX
    \item DeepWalk~\cite{perozzi2014deepwalk}, Text-Associated DeepWalk (TADW)~\cite{yang2015network}
    \item Deep Neural Network for Graph Representation (DNGR)~\cite{cao2016deep}
    \item Spectral Clustering~\cite{tang2011leveraging}
\end{itemize}
Representative GNN unsupervised methods that rely on both structure and features include GAE/VGAE, which are well-known reconstruction-based methods, and ARGVA and its variants, which additionally learn the data distribution in an adversarial fashion. GAE*/VGAE*, Spectral Clustering, DeepWalk, and DNRG rely only on graph structure. TADW is a version of DeepWalk that additionally accounts for features.

To evaluate how GIC's unsupervised representations compared to semi-supervised approaches that use the labels during representation learning, we used the following semi-supervised approaches. 
\begin{itemize}
    \item Graph Convolutional Network (GCN)~\cite{kipf2016semi}
    \item Graph Attention Network (GAT)~\cite{velickovic2018graph}
    \item Mixture Model Network (MoNet)~\cite{monti2017geometric} 
    \item GraphSAGE (GS)~\cite{hamilton2017inductive}, and its aggregator types GS-mean, GS-meanpool, GS-maxpool
\end{itemize}
which are well-known GNN-based methods.

\begin{table}
    \centering
    \resizebox{0.9\columnwidth}{!}{ 
    \begin{tabular}{lrc rrr}
    \toprule
         $\alpha$&$\beta$&$K$ & CORA & CiteSeer & PubMed \\
         \midrule
         0& 10& \text{\#classes} & 0.153 & 0.144 & 0.091\\
         0& 10& 32 & 0.206 & 0.140 & 0.094 \\
         0& 10& 128 & 0.226 & 0.165 & 0.081\\
         0& 100& \text{\#classes} & 0.213 & 0.144 & 0.086\\
         0& 100& 32 & 0.188 & 0.173 & 0.092 \\
         0& 100& 128 & 0.198 & 0.172 & 0.091\\
         \\
         0.25& 10& \text{\#classes} & 0.216 & 0.146 & 0.099\\
         0.25& 10& 32 &\textbf{0.253} & 0.157 & 0.102\\
         0.25& 10& 128 & 0.223 & 0.173 & 0.099\\
         0.25& 100& \text{\#classes} & 0.236 & 0.172 & 0.066\\
         0.25& 100& 32 & 0.239 & \textbf{0.177} & 0.101\\
         0.25& 100& 128 & 0.232 & 0.165 & 0.102\\
         \\
          &  &  &   &   &  \\
         \bottomrule
    \end{tabular}
    \begin{tabular}{lrc rrr}
    \toprule
         $\alpha$&$\beta$&$K$ & CORA & CiteSeer & PubMed \\
         \midrule
         0.5& 10& \text{\#classes} & 0.248 & 0.170 &0.110\\
         0.5& 10& 32 & 0.250 & \textbf{0.176} &0.111\\
         0.5& 10& 128 & \textbf{0.264} & 0.158 &0.113\\
         0.5& 100& \text{\#classes} & 0.224 & 0.148 &0.115\\
         0.5& 100& 32 & 0.234 & 0.153 &0.107\\
         0.5& 100& 128 & 0.245 & 0.170 &0.105\\
         \\
         0.75& 10& \text{\#classes} & 0.243 & 0.160 & \textbf{0.125}\\
         0.75& 10& 32 & 0.243 & 0.125 & \textbf{0.119}\\
         0.75& 10& 128 & 0.250 & 0.166 & 0.114\\
         0.75& 100& \text{\#classes} & 0.236 & 0.161 & 0.115\\
         0.75& 100& 32 & 0.244 & 0.165 & 0.105\\
         0.75& 100& 128 & 0.225 & 0.159 & 0.102\\
         \\
         1 & - & - & 0.191 & 0.145 & 0.077\\
         
        \bottomrule
    \end{tabular}
    }
    \caption{Silhouette scores of the t-SNE 2D projections of the learned node representations, based on different hyperparameter $\alpha, \beta, K$ values. Bold font indicates the two best scores for each dataset.}
    \label{tabablation}
\end{table}

\begin{figure}
\begin{subfigure}[b]{\linewidth}
        \begin{subfigure}[b]{.46\linewidth}
        \begin{subfigure}[b]{.32\linewidth}
        \centering
        $K=7$
        \includegraphics[width=\linewidth]{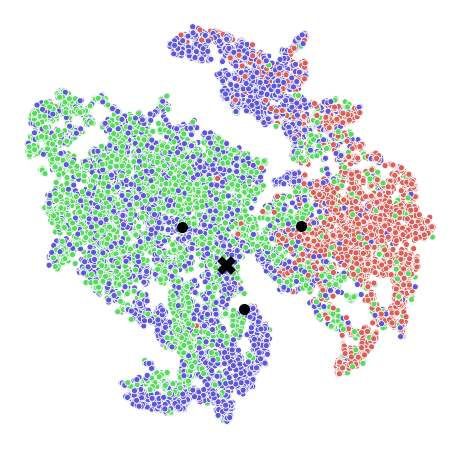}
        SIL=0.091
        \end{subfigure}
        \begin{subfigure}[b]{.32\linewidth}
        \centering
        $K=32$
        \includegraphics[width=\linewidth]{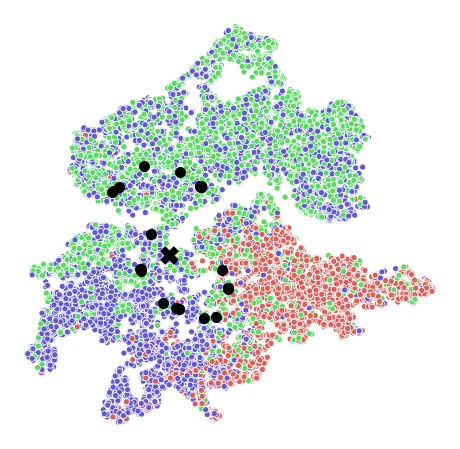}
        SIL=0.094
        \end{subfigure}
        \begin{subfigure}[b]{.32\linewidth}
        \centering
        $K=128$
        \includegraphics[width=\linewidth]{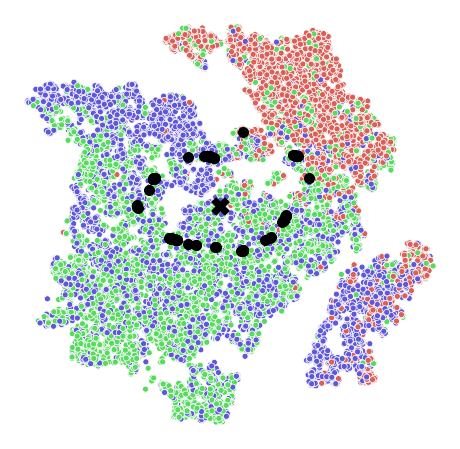}
        SIL=0.081
        \end{subfigure}
        \caption{$\alpha=0, \beta=10$}
        \label{figpa0b10}
        \end{subfigure}
        \hfill
        \begin{subfigure}[b]{.46\linewidth}
        \begin{subfigure}[b]{.32\linewidth}
        \centering
        $K=7$
        \includegraphics[width=\linewidth]{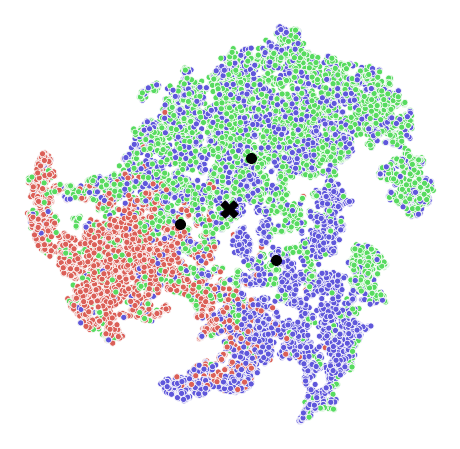}
        SIL=0.086
        \end{subfigure}
        \begin{subfigure}[b]{.32\linewidth}
        \centering
        $K=32$
        \includegraphics[width=\linewidth]{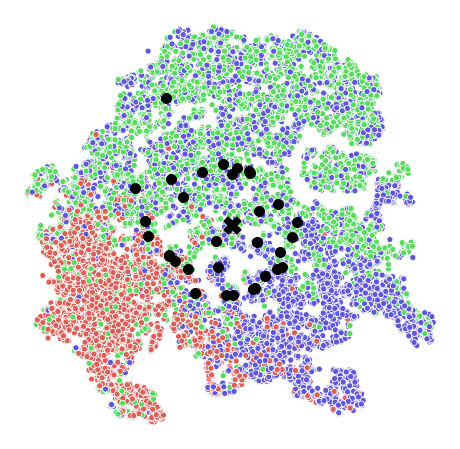}
        SIL=0.092
        \end{subfigure}
        \begin{subfigure}[b]{.32\linewidth}
        \centering
        $K=128$
        \includegraphics[width=\linewidth]{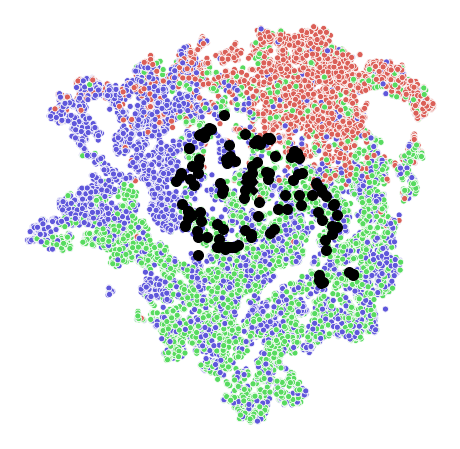}
        SIL=0.091
        \end{subfigure}
        \caption{$\alpha=0, \beta=100$}
        \label{figpa0b100}
        \end{subfigure}
\end{subfigure}
\par\bigskip
\begin{subfigure}[b]{\linewidth}
    \begin{subfigure}[b]{.46\linewidth}
        \begin{subfigure}[b]{.32\linewidth}
        \centering
        \includegraphics[width=\linewidth]{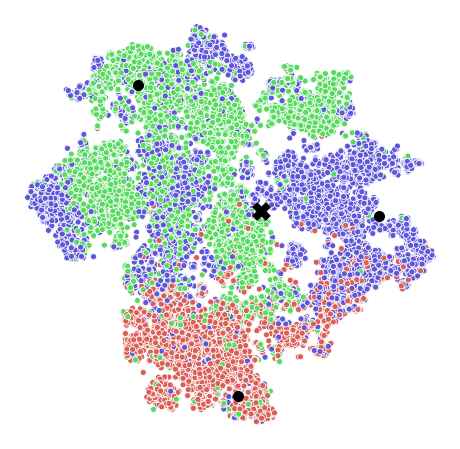}
        SIL=0.110
        \end{subfigure}
        \begin{subfigure}[b]{.32\linewidth}
        \centering
        \includegraphics[width=\linewidth]{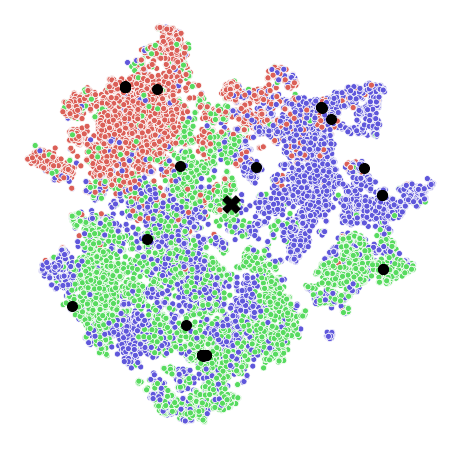}
        SIL=0.111
        \end{subfigure}
        \begin{subfigure}[b]{.32\linewidth}
        \centering
        \includegraphics[width=\linewidth]{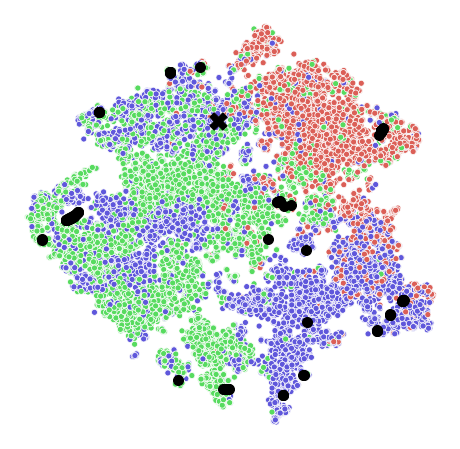}
        SIL=0.113
        \end{subfigure}
    \caption{$\alpha=0.5, \beta=10$}
    \label{figpa05b10}
    \end{subfigure}
    \hfill
    \begin{subfigure}[b]{.46\linewidth}
        \begin{subfigure}[b]{.32\linewidth}
        \centering
        \includegraphics[width=\linewidth]{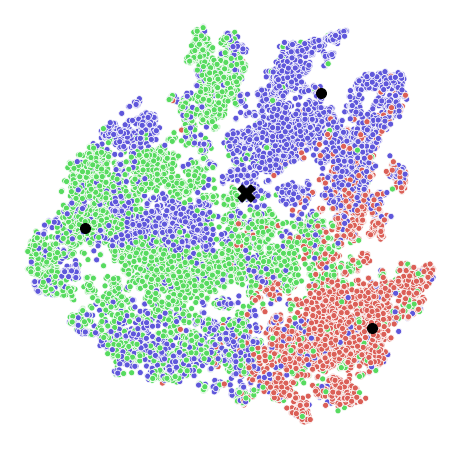}
        SIL=0.115
        \end{subfigure}
        \begin{subfigure}[b]{.32\linewidth}
        \centering
        \includegraphics[width=\linewidth]{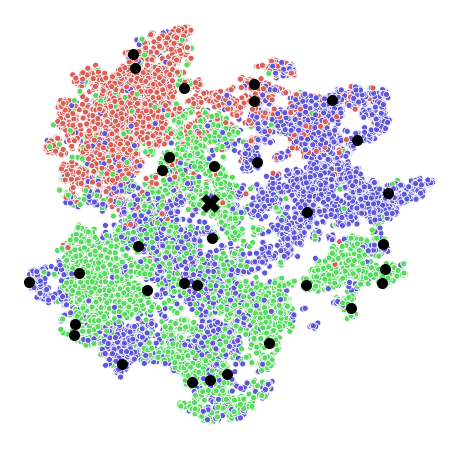}
        SIL=0.107
        \end{subfigure}
        \begin{subfigure}[b]{.32\linewidth}
        \centering
        \includegraphics[width=\linewidth]{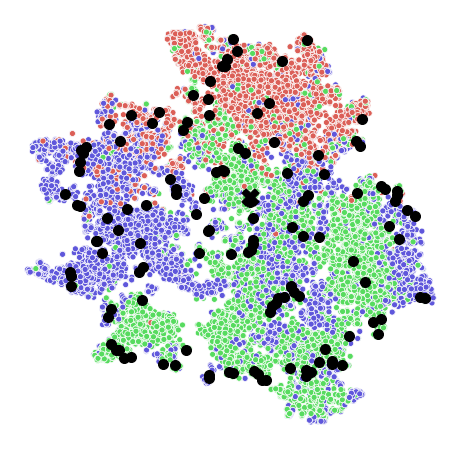}
        SIL=0.105
        \end{subfigure}
        \caption{$\alpha=0.5, \beta=100$}
        \label{figpa05b100}
        \end{subfigure}
\end{subfigure}
\caption{t-SNE plots for PubMed dataset and the corresponding silhouette scores (SIL) with $\alpha\in\{0,0.5\}, \beta \in\{10,100\}, K\in\{\#\text{classes}, 32, 128\}$. The corresponding SIL score for DGI ($\alpha=1$) is 0.077.}
\label{figpubmed}
\end{figure}
\subsection{Hardware and software}

We implemented GIC using the Deep Graph Library~\cite{wang2019dgl} and PyTorch~\cite{paszke2017automatic} (the code will be made publicly available after the paper is accepted). We also implemented GIC by modifying DGI's original implementation\footnote{DGI implementations are found in \texttt{\url{https://github.com/dmlc/dgl/tree/master/examples/pytorch/dgi}} and \texttt{\url{https://github.com/PetarV-/DGI}}.}, which we use in some experiments, e.g., link prediction and clustering. All experiments were performed on a Nvidia Geforce RTX-2070 GPU on a i5-8400 CPU and 32GB RAM machine. 

\subsection{Additional experiments}

The additional experiments are part of the Ablation Study section. In Table~\ref{tabablation}, we present silhouette scores (SIL) for different datasets, based on different hyperparameter $\alpha,\beta,K$ values. In Figure~\ref{figpubmed}, we also plot the learned node representations for PubMed dataset, after t-SNE 2D projection is applied.

\end{document}